\documentclass[runningheads]{llncs}
\usepackage{graphicx}

\usepackage{tikz}
\usepackage{comment}
\usepackage{amsmath,amssymb} 
\usepackage{color}

\usepackage{cite}
\usepackage{multirow}
\usepackage{pifont}
\usepackage{orcidlink}

\newcommand{\cmark}{\ding{51}}%
\usepackage[accsupp]{axessibility}  

\begin{document}
\pagestyle{headings}
\mainmatter
\def\ECCVSubNumber{27}  

\title{Consistency-based Self-supervised Learning for Temporal Anomaly Localization} 

\titlerunning{Consistency-based Self-supervision for Temporal Anomaly Localization}
%
\author{Aniello Panariello\inst{1}\orcidlink{0000-0002-1940-7703} \and
Angelo Porrello\inst{1}\orcidlink{0000-0002-9022-8484} \and
Simone Calderara\inst{1}\orcidlink{0000-0001-9056-1538} \and\\
Rita Cucchiara\inst{1}\orcidlink{0000-0002-2239-283X} }
\authorrunning{A. Panariello et al.}
%
\institute{AImageLab, University of Modena and Reggio Emilia, Italy
\email{firstname.lastname@unimore.it}\\
}
\maketitle

\begin{abstract}
This work tackles Weakly Supervised Anomaly detection, in which a predictor is allowed to learn not only from normal examples but also from a few labeled anomalies made available during training. In particular, we deal with the localization of anomalous activities within the video stream: this is a very challenging scenario, as training examples come only with video-level annotations (and not frame-level). Several recent works have proposed various regularization terms to address it \textit{i.e.} by enforcing sparsity and smoothness constraints over the weakly-learned frame-level anomaly scores. In this work, we get inspired by recent advances within the field of self-supervised learning and ask the model to yield the same scores for different augmentations of the same video sequence. We show that enforcing such an alignment improves the performance of the model on XD-Violence. 

\end{abstract}
\section{Introduction}
The goal of Video Anomaly Detection is to detect and localize anomalous events occurring in a video stream. Usually, these events involve human actions such as abuse, falling, fighting, theft, etc. In the last decades, such a task has gained relevance thanks to its potential for a video surveillance pipeline. In light of the widespread presence of CCTV cameras, employing enough personnel to examine all the footage is indeed infeasible and, therefore, automatic tools must be exploited.

Traditional approaches leverage low level features~\cite{dalal2005histograms} to deal with this task, computed either on visual cues~\cite{benezeth2009abnormal,kim2009observe} or object trajectories~\cite{calderara2011detecting, morris2011trajectory}. These techniques -- which could suffer and be unreliable in some contexts~\cite{hu2016video, medel2016anomaly} -- have been surpassed by reconstruction-based approaches~\cite{hasan2016learning}: these methods require a set of normal data to learn a model of regularity; afterward, an example can be assessed as "anomalous" if it substantially deviates from the learned model. In this regard, the architectural design often resorts to deep unsupervised autoencoders~\cite{zhu2019motion, wang2021robust}, which may include several strategies for regularizing the latent space.

The major recent trend~\cite{sultani2018real} regards the exploitation of weak supervision, which provides the learning phase also with examples from the anomalous classes. However, the annotations come in a video-level format: the learner does not know in which time steps the anomaly will show, but only if it does at least one time. For such a reason, the proposed techniques often fall into the framework of multiple instance learning~\cite{tian2021weakly, wu2020not, feng2021mist}, devising additional optimization constraints that mitigate the lack of frame-level annotations.

In this work, we attempt to complement the existing regularization techniques with an idea from the fields of self-supervised learning and consistency regularization~\cite{sohn2020fixmatch, xie2020unsupervised,chen2020simple,boschini2021continual}. In this context, data augmentation is exploited to generate positive pairs, consisting of a couple of slightly different versions of the same example. Afterward, the network is trained to output very similar representations, in a self-supervised manner (no class labels are required). Similarly, we devise a data augmentation strategy tailored to sequences and encourage the network to assign the same frame-level anomaly scores for the elements of each positive pair. Such a strategy resembles the smoothness constraint often imposed in recent works; however, while those approaches focus on adjacent time steps, our approach can randomly span over longer temporal windows.

To evaluate the merits of our proposal, we conduct experiments on the XD-Violence~\cite{wu2020not} dataset. We found that the presence of our regularization term yields remarkable improvements, although being not yet enough to reach state-of-the-art approaches.
\section{Related work}
\subsubsection{Video Anomaly Detection}
Traditionally researchers and practitioners exploit object trajectories~\cite{calderara2011detecting, morris2011trajectory,monti2022many}, low-level handcrafted features~\cite{benezeth2009abnormal, kim2009observe, mehran2009abnormal} (such as Histogram of Gradients and Histogram of Flows), and connected component analysis cues~\cite{amraee2018anomaly,bolelli2019spaghetti,bolelli2021one}. These traditional approaches have proven to be effective on benchmark datasets, but turn out to be still ineffective when used on a real domain. This means that these methods do not adapt well to anomalies that have never been seen before. For this reason, the task of anomaly detection has recently been approached with deep neural networks.

Most recent works lean towards unsupervised methods~\cite{hasan2016learning,abati2019latent} usually based on deep autoencoders. These models leverage the reconstruction error to measure how much an incoming example conforms to the training set distribution. Remarkably, the authors of~\cite{liu2018future} resorted to a variation of this common paradigm: namely, they proposed an approach that learns the distribution of normal data by guessing the appearance of future frames. In this respect, the idea was to compare the actual frame and the predicted one: if their difference is high, then it is possible to assume that an anomalous event occurred.

While it is easy to label whole videos (\textit{e.g.}, anomaly present or not), the availability of fine-grained labeled anomalous videos is often scarce. For this reason, weakly supervised methods have seen a great advance. Among these works, Sultani \textit{et al.}~\cite{sultani2018real} introduced a new pattern for video anomaly detection termed Multiple Instance Learning (MIL), upon which most of the subsequent weakly supervised methods developed. In this scenario, a positive video (\textit{i.e.} containing an anomaly) and a negative one are taken into account at each iteration. These videos are usually split into segments; the model assigns them an anomaly score through a sequence of convolutional layers followed by an MLP. The scores are collected into a positive bag and a negative bag: as in the former, there is at least an anomaly (while in the latter there are no anomalies) the minimum score from the positive bag has to be as far as possible from the maximum score of the negative bag. To impose this constraint, the objective function contains a MIL ranking loss as well as smoothness and sparsity constraints. Zhou \textit{et al.}~\cite{zhu2019motion} introduced the attention mechanism combined with the MIL approach, to improve the localization of the anomalies.
\subsubsection{Temporal Action Localization}
The majority of the works dealing with Temporal Action Localization are either fully supervised or weakly supervised. The former ones can be broadly categorized into one stage~\cite{long2019gaussian, lin2017single} and two stages methods~\cite{xu2017r, zeng2019graph, lin2019bmn}. For the first type, in works such as~\cite{lin2017single}, action boundaries and labels are predicted simultaneously; in~\cite{long2019gaussian} this concept is developed by exploiting Gaussian kernels to dynamically optimize the temporal scale of each action proposal. On the other hand, two-stage methods initially generate temporal action proposals and then classify them. A manner to produce proposals is by exploiting the anchor mechanism~\cite{chao2018rethinking, gao2017turn, yang2020revisiting}, sliding window~\cite{shou2016temporal} or combining confident starting and ending frames of an action~\cite{lin2018bsn, lin2019bmn}.

Weakly supervised methods have been pioneered by UntrimmedNet~\cite{wang2017untrimmednets}, STPN~\cite{nguyen2018weakly} and AutoLoc~\cite{shou2018autoloc}, in which the action instances are localized by applying a threshold on the class activation sequence. This paradigm has been recently brought into the video anomaly detection settings in~\cite{wu2021weakly, lv2021localizing}.
\section{Proposed Method}
In the following, we present the proposed model and its training objective. The third part is dedicated to explaining the process of proposal generation, which consists of a post-processing step grouping adjacent similar scores into contiguous discrete intervals. 
\subsection{Model}\label{sec:model}
\begin{figure}[t]
    \centering
    \includegraphics[width=\textwidth]{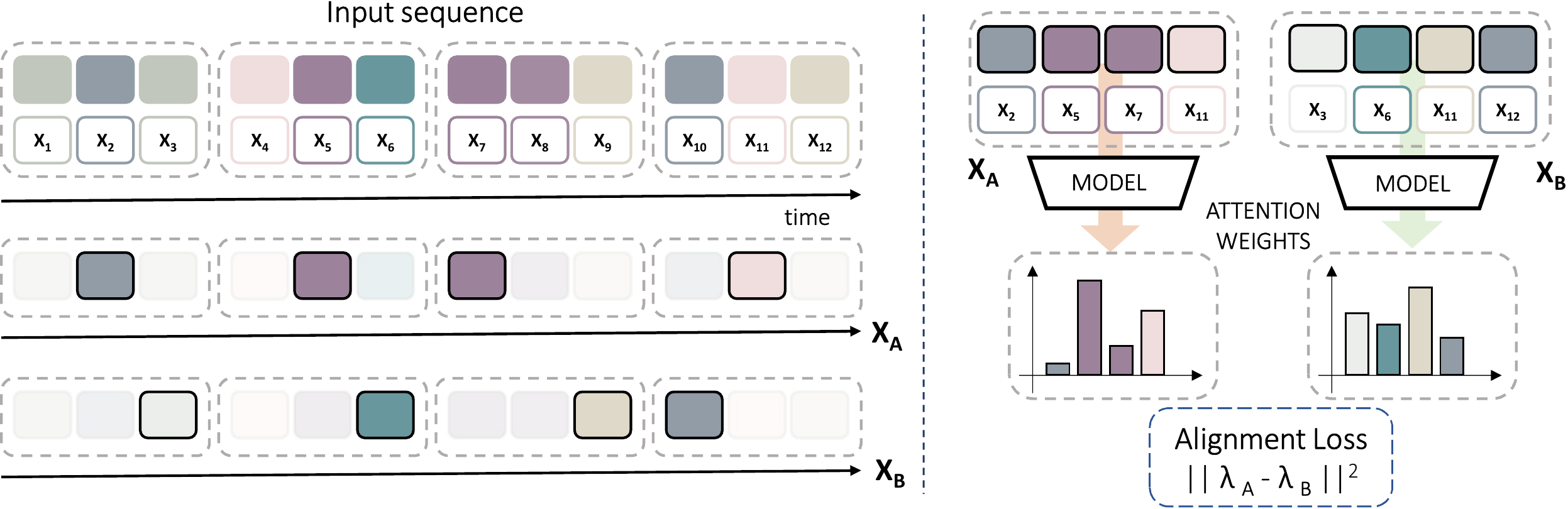}
    \caption{Overview of the proposed framework. (left) Augmentation function sampling two slightly different sequences out of a single one. The original sequence gets split into windows and for each, we randomly sample a single feature vector. This is done twice to obtain two sequences $X^A$ and $X^B$. (right) Both sequences are separately fed to the model, obtaining two sequences of attention weights $\lambda_A$ and $\lambda_B$, pulled closer together by the alignment loss.}
    \label{fig:model}
\end{figure}
In our set-up, each video is split into segments of 16 frames, with no overlap between consecutive segments. To extract video-level features, some works~\cite{yue2015beyond,cascianelli2019role} opted for combining 2D-CNNs and Recurrent Neural Networks; instead, we give each segment to a pre-trained I3D network~\cite{carreira2017quo}. Indeed, the authors of~\cite{koshti2020video} have shown that the I3D features can prove to be effective for video-anomaly detection, even when a shallow classifier as XGBoost~\cite{chen2016xgboost,candeloro2020predicting} is employed for later classification. In our case, the I3D network is pre-trained on the Kinetics~\cite{kay2017kinetics} dataset and not fine-tuned on our target data. 

This way, each example is represented by a variable-length sequence of $T$ feature vectors $\mathcal{X} = (x_1, x_2, \dots, x_T)$. During training, examples come with a label $y$ indicating whether an anomalous event appears in that sequence at least one time. Hence, given a training set of examples $\{(\mathcal{X}_j,y_j)\}_{j = 1}^{\mathcal{N}}$ forged in the above-mentioned manner, we seek to train a neural network $f(\cdot; \theta)$ that solves such a task with the lowest empirical error:
\begin{equation}
\min_{\theta} \quad \frac{1}{\mathcal{N}}\sum_{i=1}^{N}{\operatorname{BCE}(f(\mathcal{X}_i; \theta), y_i)}
\end{equation}
where $\operatorname{BCE}(\cdot, \cdot)$ stands for the binary cross entropy loss function.
For the architectural design of $f(\cdot; \theta)$, we took inspiration from~\cite{nguyen2018weakly}. It consists of two main parts, discussed in the following paragraphs: namely, the computation of attention coefficients and the creation of an aggregate video-level representation. 
\subsubsection{Attention coefficients} The aim of this module is to assign a weight $\lambda_t \in [0,1]$ to each element of the input sequence. As explained in the following, these weights will identify the most salient segments of the video \textit{i.e.}, the likelihood of having observed an abnormal event within each segment. The module initially performs a masked temporal 1D convolution~\cite{lea2017temporal} to allow each feature vector to encode information from the past. Such a transformation -- which does not alter the number of input feature vectors -- is followed by two fully connected layers activated by ReLU functions, except for the last layer where a sigmoid function is employed.
\subsubsection{Video-level representation} Once we have guessed the attention values, we exploit them to aggregate the input feature vectors. Such an operation -- which resembles a temporal weighted average pooling presented in~\cite{porrello2019spotting} -- produces a single feature vector with unchanged dimensionality; formally:
\begin{equation}
    \mathbf{x} = \sum_{t=1}^{T} \lambda_t x_t.
\end{equation}
We finally feed it to a classifier $g(\cdot)$, composed of two fully connected layers. The final output represents the guess of the network for the value of $y$.
\subsection{Training Objective} As mentioned before, we train our network in a weakly supervised fashion \textit{i.e.}, only video level labels are provided to the learner. However, to provide a stronger training signal and to encourage attention coefficients to highlight salient events, we follow recent works~\cite{cai2021appearance, yu2020cloze} and use some additional regularization terms that encode a prior knowledge we retain about the dynamics of abnormality.

Often, the presence of anomalous activities is characterized by the properties of sparseness and smoothness. Namely, anomalies appear rarely (\textit{i.e.}, normal events dominate) and transitions between the two modalities usually occur throughout multiple frames. Such a peculiarity represents a prior we would like to enforce over the scores learned by the model: the majority of them should be close to zero and have similar values for video segments. In formal terms, the first constraint can be injected by penalizing the $l_1$ norm~\cite{cascianelli2021data} of the attention weights, as follows:
\begin{equation}
    \mathcal{L}_{sp} = \Vert \mathbf{\lambda} \Vert_1,
\end{equation}
while the second one can be carried out by imposing adjacent coefficients to vary as little as possible:
\begin{equation}
    \mathcal{L}_{sm} = \sum_{t=1}^{T-1} (\lambda_t - \lambda_{t+1})^2.
\end{equation}
\subsubsection{Alignment loss} Our main contribution consists in adding a consistency-based regularization term to the overall objective function. Overall, the idea is to generate two slightly different sequences out of a single one and, then, to encourage the model to produce the same attention coefficients for the two inputs.

To do so, we introduce a data augmentation function, shown in Fig.~\ref{fig:model}, that allows us to forge different versions $\mathcal{X}_A$ and $\mathcal{X}_B$ from the same example $\mathcal{X}$. In more detail, we split each sequence $(x_1, x_2, \dots, x_T)$ into fixed-size blocks, whose length $L$ is a hyperparameter we always set to $3$; afterward, we randomly choose a feature vector within each block.

Once the variants $\mathcal{X}_A$ and $\mathcal{X}_B$ have been created, we ask the network to minimize the following objective function:
\begin{equation}
    \mathcal{L}_{a} = \sum_{t=1}^{T} (\lambda^A_t - \lambda^B_{t})^2,
\end{equation}
where $\lambda^A$ and $\lambda^B$ are respectively the attention coefficients computed by the network for $\mathcal{X}_A$ and $\mathcal{X}_B$. With this additional regularization term, we seek to enforce that not only adjacent time-steps should have the same weight, but also those lying within a wider temporal horizon. 
\subsubsection{Overall objective} Finally, the objective function will be:
\begin{equation}
    \mathcal{L} \equiv \mathcal{L}_{cl} + \alpha \mathcal{L}_{sp} + \beta \mathcal{L}_{sm} + \gamma \mathcal{L}_{a},
\end{equation}
where the parameters $\alpha$, $\beta$, and $\gamma$ give different weights to each loss component and $ \mathcal{L}_{cl}$ is the binary cross entropy loss function.
\subsection{Temporal Proposal}
During inference, we refine the anomaly scores by applying a post-processing step. Usually, two segments considered paramount by the network are interleaved by ``holes'', mostly due to noisy acquisitions or poor representations. The purpose of this phase is therefore to merge temporally close detections in a single retrieved candidate. In particular, as done in~\cite{zhou2016learning}, we initially take out from the candidate set all those time-steps whose corresponding attention scores are lower than a certain threshold (in our experiments, $< 0.35$). The remaining non-zero scores will be used to generate the temporal proposal. 

To generate the proposals, we do not use the rough coefficients, but instead a more refined version. In particular, we compute a 1-d activation map in the temporal domain, called Temporal Class Activation Map (T-CAM)~\cite{zhou2016learning}, which indicates the relevance of the segment $t$ in the prediction of one of the two classes involved (\textit{normal vs anomalous}). Each value $a_t$ of such activation map is computed as $a_t = g(x_t)$, \textit{i.e.}, the guess of the classifier $g(\cdot)$ (introduced in Sec.~\ref{sec:model}) if masking the contributions of all time-steps except the $t$-th one. Furthermore, we extract the Weighted T-CAM, which combines the attention weight and the T-CAM activation values, \textit{i.e.}, $\psi_t = \lambda_t \cdot a_t$. This operation let us emphasize the most important features for generating the proposal.

The last operation involves interpolating the weighted scores in the temporal axis and taking the bounding box that covers the largest connected component~\cite{nguyen2018weakly} to generate the final proposal~\cite{zhou2016learning}. The anomaly score for each proposal is then computed as:
\begin{equation}
    \sum_{t=t_{start}}^{t_{end}} = \lambda_t \cdot \frac{a_t}{t_{end}-t_{start}-1},
\end{equation}
where $t_{start}$ and $t_{end}$ represent the beginning and the ending of a single proposal.
\section{Experiments}
\begin{table}[t]
    \centering
    \begin{tabular}{c|c|c|c|c|c|c|c|c}
                    & \multicolumn{2}{c|}{Video Level} & \multicolumn{2}{c|}{Segment Level} & \multicolumn{2}{c|}{Frame Level Proposal} & \multicolumn{2}{c}{Frame Level} \\ \hline
    Align Loss  &  AUC\%    & AP\%  & AUC\% & AP\%  & AUC\% & AP\%  & AUC\% & AP\%  \\
        -       & \textbf{97.91} & \textbf{98.36} & 84.39 & 66.75 & 85.14 & 68.01 & 84.57 & 65.96 \\ 
    \cmark      & 97.79     & 98.28 & \textbf{85.49} &\textbf{ 66.87} & \textbf{90.23} & \textbf{71.68} & \textbf{85.65} & \textbf{66.05} \\
    \end{tabular}
    \caption{For different levels and metrics, results of our model with and without the proposed align loss. There is an improvement on almost all metrics when leveraging the proposed term.}
    \label{tab:align-ablation}
\end{table}
\begin{table}[t]
    \centering
    \begin{tabular}{c|l|c}
    Supervision & Method & AP\%    \\ \hline
    \multirow{3}{*}{\textbf{Unsupervised}} & SVM baseline  & 50.78   \\
        & OCSVM~\cite{scholkopf1999support}     & 27.25 \\
        & Hasan et al.~\cite{hasan2016learning} & 30.77 \\ \hline
    \multirow{5}{*}{\textbf{Weakly Supervised}} 
        & Ours (no align loss)                  & 68.01 \\
        & Ours                                  & 71.68 \\\cline{2-3}
        & Sultani et al.~\cite{sultani2018real} & 75.68  \\
        & Wu et al.~\cite{wu2020not}            & 75.41 \\
        & RTFM~\cite{tian2021weakly}            & 77.81 \\
    \end{tabular}
    \caption{Comparison with recent works. We report the frame-level AP score on XD-Violence for both unsupervised and weakly-supervised methods. All the competitors exploit the I3D network for extracting features from RGB frames.}
    \label{tab:competitors}
\end{table}
We conduct our experiments on the \textbf{XD-Violence Dataset}~\cite{wu2020not}, a multi-modal dataset that contains scenes from different sources such as movies, sports, games, news, and live scenes. It holds a great variety concerning the devices used for video acquisition; indeed, the examples were captured by CCTV cameras, hand-held cameras, or car driving recorders. There are $4754$ videos for a total of $217$ hours: among all these, $2405$ are violent videos and $2349$ are non-violent. The training set consists of $3954$ examples; the test set, instead, features $800$ ones, split into $500$ violent and $300$ non-violent videos. The dataset comes with multiple modalities such as RGB, optical flow, and audio; however, we restrict our analysis only to the RGB input domain.

XD-Violence also comes with segment-level ground truth labels; however, we only use video-level annotation during training, conforming to the weakly supervised setting. Differently, we exploit both the segment-level and frame-level annotations during the test phase, thus evaluating the model's capabilities for fine-grained localization.
\subsubsection{Metrics}
We used the most popular metrics in anomaly detection settings, namely the Area Under Receiver Operating Characteristic Curve (AUC) and the Area Under Precision-Recall Curve (AP). While the AUC tends to be optimistic in the case of unbalanced datasets, the AP gives a more accurate evaluation of these scenarios.

We assess the performance of the classification head by providing these two metrics at the video level; therefore, we use the attention scores, their interpolation, as well as the temporal proposals to assess the other grains.
\begin{figure}[t]
    \centering
    \includegraphics[width=\textwidth]{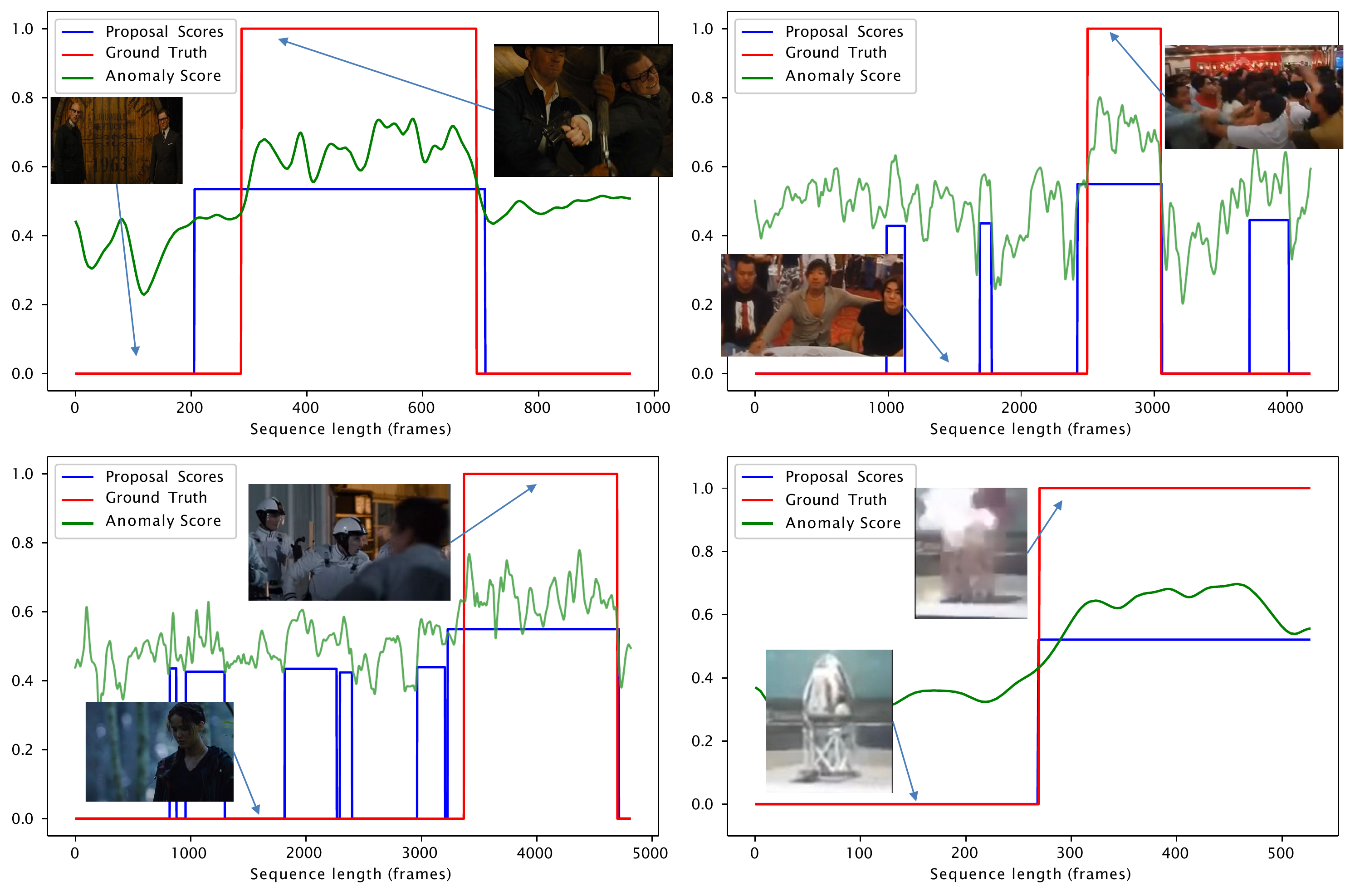}
    \caption{Qualitative examples of the capabilities of our model to perform anomaly localization. The temporal proposal scores are indicated with a blue line, while the weighted T-CAM scores and the ground truth are shown in green and red respectively.}
    \label{fig:align}
\end{figure}
\subsubsection{Training Details}
We use the Adam optimizer~\cite{kingma2014adam} with a learning rate of $10^{-4}$ for the first 10 epochs and $10^{-5}$ for the remaining 40 epochs. The hyper-parameters for the loss components are set to: $\alpha=2\times 10^{-8}$, $\beta=0.002$, $\gamma=0.5$. The threshold for taking out the low weighted T-CAM scores is set to $0.35$; the batch size equals 8.
\subsubsection{Results}
Tab~\ref{tab:align-ablation} reports the comparison between the baseline approach with and without the proposed alignment objective. It can be observed that its addition leads to a remarkable improvement in most of the metrics. In particular, we have a gain of about $1\%$ in AUC for segment and frame level metrics, while the AP remains almost the same. The greatest improvement subsists for the temporal proposal metric, where we gain 5 points in AUC and around 4 points in AP. The video level metrics remain approximately the same but still very high.

When comparing our approach with other recent works (see Tab~\ref{tab:competitors}), it can be seen that it outperforms the unsupervised state-of-the-art methods; however, it is in turn surpassed by the weakly supervised ones. We conjecture that such a gap is mainly due to the bag representations inherent in these approaches, which could confer superior robustness; therefore, we leave to future works the extension of our idea to these methods.
\subsubsection{Qualitative analysis} Fig~\ref{fig:align} presents several qualitative results, showing two fight scenes in the first row, and a riot and an explosion in the second one. We notice that the scores rightly increase when an anomalous action begins. Unfortunately, they remain always close to the uncertainty regime (assuming a score around $0.5$) and never tend towards discrete decisions. In future works, we are going to address also this issue.

Looking at the original videos, we could also explain why the scores yielded by the model are noisy and subject to local fluctuations. Indeed, as sudden characters or camera' movements are likely to intervene, they are mistaken for real anomalies by the model, which is susceptible to these visual discontinuities due to the lack of segment-level annotations during training. Differently, when relying on the entire video sequence, the model can conversely recognize them correctly.
\section{Conclusions}
This work proposes a novel strategy to learn effective frame-level scores in weakly supervised settings when only video-level annotations are made available to the learner. Our proposal -- which builds upon recent advances in the field of self-supervised learning -- relies on maximizing the alignment between the attention weights of two different augmentations of the same input sequence. We show that a base network equipped also with other common regularization strategies (\textit{e.g.} sparsity and smoothness) brings even more improvements. We found that it is not enough to achieve state-of-the-art performance; however, we leave aside for future works a comprehensive investigation of its applicability to more advanced architectures.

\subsubsection{Acknowledgments.} This work has been supported in part by the InSecTT project, funded by the Electronic Component Systems for European Leadership Joint Undertaking under grant agreement 876038. The Joint Undertaking receives support from the European Union’s Horizon 2020 research and innovation programme and AU, SWE, SPA, IT, FR, POR, IRE, FIN, SLO, PO, NED and TUR. The document reflects only the author’s view and the Commission is not responsible for any use that may be made of the information it contains.

%
%
\bibliographystyle{splncs04}
\bibliography{egbib}

\begin{thebibliography}{10}
\providecommand{\url}[1]{\texttt{#1}}
\providecommand{\urlprefix}{URL }
\providecommand{\doi}[1]{https://doi.org/#1}

\bibitem{abati2019latent}
Abati, D., Porrello, A., Calderara, S., Cucchiara, R.: Latent space
  autoregression for novelty detection. In: Proceedings of the IEEE conference
  on Computer Vision and Pattern Recognition (2019)

\bibitem{amraee2018anomaly}
Amraee, S., Vafaei, A., Jamshidi, K., Adibi, P.: Anomaly detection and
  localization in crowded scenes using connected component analysis. Multimedia
  Tools and Applications  (2018)

\bibitem{benezeth2009abnormal}
Benezeth, Y., Jodoin, P.M., Saligrama, V., Rosenberger, C.: Abnormal events
  detection based on spatio-temporal co-occurences. In: Proceedings of the IEEE
  conference on Computer Vision and Pattern Recognition. IEEE (2009)

\bibitem{bolelli2019spaghetti}
Bolelli, F., Allegretti, S., Baraldi, L., Grana, C.: Spaghetti labeling:
  Directed acyclic graphs for block-based connected components labeling. IEEE
  Transactions on Image Processing  (2019)

\bibitem{bolelli2021one}
Bolelli, F., Allegretti, S., Grana, C.: One dag to rule them all. IEEE
  Transactions on Pattern Analysis and Machine Intelligence  (2021)

\bibitem{boschini2021continual}
Boschini, M., Buzzega, P., Bonicelli, L., Porrello, A., Calderara, S.:
  Continual semi-supervised learning through contrastive interpolation
  consistency. arXiv preprint arXiv:2108.06552  (2021)

\bibitem{cai2021appearance}
Cai, R., Zhang, H., Liu, W., Gao, S., Hao, Z.: Appearance-motion memory
  consistency network for video anomaly detection. In: Proceedings of the AAAI
  Conference on Artificial Intelligence (2021)

\bibitem{calderara2011detecting}
Calderara, S., Heinemann, U., Prati, A., Cucchiara, R., Tishby, N.: Detecting
  anomalies in people’s trajectories using spectral graph analysis. Computer
  Vision and Image Understanding  (2011)

\bibitem{candeloro2020predicting}
Candeloro, L., Ippoliti, C., Iapaolo, F., Monaco, F., Morelli, D., Cuccu, R.,
  Fronte, P., Calderara, S., Vincenzi, S., Porrello, A., et~al.: Predicting wnv
  circulation in italy using earth observation data and extreme gradient
  boosting model. Remote Sensing  (2020)

\bibitem{carreira2017quo}
Carreira, J., Zisserman, A.: Quo vadis, action recognition? a new model and the
  kinetics dataset. In: Proceedings of the IEEE conference on Computer Vision
  and Pattern Recognition (2017)

\bibitem{cascianelli2021data}
Cascianelli, S., Costante, G., Crocetti, F., Ricci, E., Valigi, P.,
  Luca~Fravolini, M.: Data-based design of robust fault detection and isolation
  residuals via lasso optimization and bayesian filtering. Asian Journal of
  Control  (2021)

\bibitem{cascianelli2019role}
Cascianelli, S., Costante, G., Devo, A., Ciarfuglia, T.A., Valigi, P.,
  Fravolini, M.L.: The role of the input in natural language video description.
  IEEE Transactions on Multimedia  (2019)

\bibitem{chao2018rethinking}
Chao, Y.W., Vijayanarasimhan, S., Seybold, B., Ross, D.A., Deng, J.,
  Sukthankar, R.: Rethinking the faster r-cnn architecture for temporal action
  localization. In: Proceedings of the IEEE conference on Computer Vision and
  Pattern Recognition (2018)

\bibitem{chen2016xgboost}
Chen, T., Guestrin, C.: Xgboost: A scalable tree boosting system. In:
  Proceedings of the 22nd acm sigkdd international conference on knowledge
  discovery and data mining (2016)

\bibitem{chen2020simple}
Chen, T., Kornblith, S., Norouzi, M., Hinton, G.: A simple framework for
  contrastive learning of visual representations. In: International Conference
  on Machine Learning (2020)

\bibitem{dalal2005histograms}
Dalal, N., Triggs, B.: Histograms of oriented gradients for human detection.
  In: Proceedings of the IEEE conference on Computer Vision and Pattern
  Recognition. Ieee (2005)

\bibitem{feng2021mist}
Feng, J.C., Hong, F.T., Zheng, W.S.: Mist: Multiple instance self-training
  framework for video anomaly detection. In: Proceedings of the IEEE conference
  on Computer Vision and Pattern Recognition (2021)

\bibitem{gao2017turn}
Gao, J., Yang, Z., Chen, K., Sun, C., Nevatia, R.: Turn tap: Temporal unit
  regression network for temporal action proposals. In: IEEE International
  Conference on Computer Vision (2017)

\bibitem{hasan2016learning}
Hasan, M., Choi, J., Neumann, J., Roy-Chowdhury, A.K., Davis, L.S.: Learning
  temporal regularity in video sequences. In: Proceedings of the IEEE
  conference on Computer Vision and Pattern Recognition (2016)

\bibitem{hu2016video}
Hu, X., Hu, S., Huang, Y., Zhang, H., Wu, H.: Video anomaly detection using
  deep incremental slow feature analysis network. IET Computer Vision  (2016)

\bibitem{kay2017kinetics}
Kay, W., Carreira, J., Simonyan, K., Zhang, B., Hillier, C., Vijayanarasimhan,
  S., Viola, F., Green, T., Back, T., Natsev, P., et~al.: The kinetics human
  action video dataset. arXiv preprint arXiv:1705.06950  (2017)

\bibitem{kim2009observe}
Kim, J., Grauman, K.: Observe locally, infer globally: a space-time mrf for
  detecting abnormal activities with incremental updates. In: Proceedings of
  the IEEE conference on Computer Vision and Pattern Recognition. IEEE (2009)

\bibitem{kingma2014adam}
Kingma, D.P., Ba, J.: Adam: A method for stochastic optimization. In:
  International Conference on Learning Representations (2014)

\bibitem{koshti2020video}
Koshti, D., Kamoji, S., Kalnad, N., Sreekumar, S., Bhujbal, S.: Video anomaly
  detection using inflated 3d convolution network. In: International Conference
  on Inventive Computation Technologies (ICICT). IEEE (2020)

\bibitem{lea2017temporal}
Lea, C., Flynn, M.D., Vidal, R., Reiter, A., Hager, G.D.: Temporal
  convolutional networks for action segmentation and detection. In: Proceedings
  of the IEEE conference on Computer Vision and Pattern Recognition (2017)

\bibitem{lin2019bmn}
Lin, T., Liu, X., Li, X., Ding, E., Wen, S.: Bmn: Boundary-matching network for
  temporal action proposal generation. In: IEEE International Conference on
  Computer Vision (2019)

\bibitem{lin2017single}
Lin, T., Zhao, X., Shou, Z.: Single shot temporal action detection. In:
  Proceedings of the 25th ACM international conference on Multimedia (2017)

\bibitem{lin2018bsn}
Lin, T., Zhao, X., Su, H., Wang, C., Yang, M.: Bsn: Boundary sensitive network
  for temporal action proposal generation. In: Proceedings of the European
  Conference on Computer Vision (2018)

\bibitem{liu2018future}
Liu, W., Luo, W., Lian, D., Gao, S.: Future frame prediction for anomaly
  detection--a new baseline. In: Proceedings of the IEEE conference on Computer
  Vision and Pattern Recognition (2018)

\bibitem{long2019gaussian}
Long, F., Yao, T., Qiu, Z., Tian, X., Luo, J., Mei, T.: Gaussian temporal
  awareness networks for action localization. In: Proceedings of the IEEE
  conference on Computer Vision and Pattern Recognition (2019)

\bibitem{lv2021localizing}
Lv, H., Zhou, C., Cui, Z., Xu, C., Li, Y., Yang, J.: Localizing anomalies from
  weakly-labeled videos. IEEE transactions on image processing  (2021)

\bibitem{medel2016anomaly}
Medel, J.R., Savakis, A.: Anomaly detection in video using predictive
  convolutional long short-term memory networks. arXiv preprint
  arXiv:1612.00390  (2016)

\bibitem{mehran2009abnormal}
Mehran, R., Oyama, A., Shah, M.: Abnormal crowd behavior detection using social
  force model. In: Proceedings of the IEEE conference on Computer Vision and
  Pattern Recognition. IEEE (2009)

\bibitem{monti2022many}
Monti, A., Porrello, A., Calderara, S., Coscia, P., Ballan, L., Cucchiara, R.:
  How many observations are enough? knowledge distillation for trajectory
  forecasting. In: Proceedings of the IEEE conference on Computer Vision and
  Pattern Recognition (2022)

\bibitem{morris2011trajectory}
Morris, B.T., Trivedi, M.M.: Trajectory learning for activity understanding:
  Unsupervised, multilevel, and long-term adaptive approach. IEEE Transactions
  on Pattern Analysis and Machine Intelligence  (2011)

\bibitem{nguyen2018weakly}
Nguyen, P., Liu, T., Prasad, G., Han, B.: Weakly supervised action localization
  by sparse temporal pooling network. In: Proceedings of the IEEE conference on
  Computer Vision and Pattern Recognition (2018)

\bibitem{porrello2019spotting}
Porrello, A., Vincenzi, S., Buzzega, P., Calderara, S., Conte, A., Ippoliti,
  C., Candeloro, L., Di~Lorenzo, A., Dondona, A.C.: Spotting insects from
  satellites: modeling the presence of culicoides imicola through deep cnns.
  In: 2019 15th International Conference on Signal-Image Technology \&
  Internet-Based Systems (SITIS). IEEE (2019)

\bibitem{scholkopf1999support}
Sch{\"o}lkopf, B., Williamson, R.C., Smola, A., Shawe-Taylor, J., Platt, J.:
  Support vector method for novelty detection. In: Advances in Neural
  Information Processing Systems (1999)

\bibitem{shou2018autoloc}
Shou, Z., Gao, H., Zhang, L., Miyazawa, K., Chang, S.F.: Autoloc:
  Weakly-supervised temporal action localization in untrimmed videos. In:
  Proceedings of the European Conference on Computer Vision (2018)

\bibitem{shou2016temporal}
Shou, Z., Wang, D., Chang, S.F.: Temporal action localization in untrimmed
  videos via multi-stage cnns. In: Proceedings of the IEEE conference on
  Computer Vision and Pattern Recognition (2016)

\bibitem{sohn2020fixmatch}
Sohn, K., Berthelot, D., Carlini, N., Zhang, Z., Zhang, H., Raffel, C.A.,
  Cubuk, E.D., Kurakin, A., Li, C.L.: Fixmatch: Simplifying semi-supervised
  learning with consistency and confidence. Advances in Neural Information
  Processing Systems  (2020)

\bibitem{sultani2018real}
Sultani, W., Chen, C., Shah, M.: Real-world anomaly detection in surveillance
  videos. In: Proceedings of the IEEE conference on Computer Vision and Pattern
  Recognition (2018)

\bibitem{tian2021weakly}
Tian, Y., Pang, G., Chen, Y., Singh, R., Verjans, J.W., Carneiro, G.:
  Weakly-supervised video anomaly detection with robust temporal feature
  magnitude learning. In: IEEE International Conference on Computer Vision
  (2021)

\bibitem{wang2017untrimmednets}
Wang, L., Xiong, Y., Lin, D., Van~Gool, L.: Untrimmednets for weakly supervised
  action recognition and detection. In: Proceedings of the IEEE conference on
  Computer Vision and Pattern Recognition (2017)

\bibitem{wang2021robust}
Wang, X., Che, Z., Jiang, B., Xiao, N., Yang, K., Tang, J., Ye, J., Wang, J.,
  Qi, Q.: Robust unsupervised video anomaly detection by multipath frame
  prediction. IEEE transactions on neural networks and learning systems  (2021)

\bibitem{wu2021weakly}
Wu, J., Zhang, W., Li, G., Wu, W., Tan, X., Li, Y., Ding, E., Lin, L.:
  Weakly-supervised spatio-temporal anomaly detection in surveillance video.
  In: International Joint Conferences on Artificial Intelligence (2021)

\bibitem{wu2020not}
Wu, P., Liu, J., Shi, Y., Sun, Y., Shao, F., Wu, Z., Yang, Z.: Not only look,
  but also listen: Learning multimodal violence detection under weak
  supervision. In: Proceedings of the European Conference on Computer Vision
  (2020)

\bibitem{xie2020unsupervised}
Xie, Q., Dai, Z., Hovy, E., Luong, T., Le, Q.: Unsupervised data augmentation
  for consistency training. Advances in Neural Information Processing Systems
  (2020)

\bibitem{xu2017r}
Xu, H., Das, A., Saenko, K.: R-c3d: Region convolutional 3d network for
  temporal activity detection. In: IEEE International Conference on Computer
  Vision (2017)

\bibitem{yang2020revisiting}
Yang, L., Peng, H., Zhang, D., Fu, J., Han, J.: Revisiting anchor mechanisms
  for temporal action localization. IEEE Transactions on Image Processing
  (2020)

\bibitem{yu2020cloze}
Yu, G., Wang, S., Cai, Z., Zhu, E., Xu, C., Yin, J., Kloft, M.: Cloze test
  helps: Effective video anomaly detection via learning to complete video
  events. In: Proceedings of the 28th ACM International Conference on
  Multimedia. pp. 583--591 (2020)

\bibitem{yue2015beyond}
Yue-Hei~Ng, J., Hausknecht, M., Vijayanarasimhan, S., Vinyals, O., Monga, R.,
  Toderici, G.: Beyond short snippets: Deep networks for video classification.
  In: Proceedings of the IEEE conference on Computer Vision and Pattern
  Recognition (2015)

\bibitem{zeng2019graph}
Zeng, R., Huang, W., Tan, M., Rong, Y., Zhao, P., Huang, J., Gan, C.: Graph
  convolutional networks for temporal action localization. In: IEEE
  International Conference on Computer Vision (2019)

\bibitem{zhou2016learning}
Zhou, B., Khosla, A., Lapedriza, A., Oliva, A., Torralba, A.: Learning deep
  features for discriminative localization. In: Proceedings of the IEEE
  conference on Computer Vision and Pattern Recognition (2016)

\bibitem{zhu2019motion}
Zhu, Y., Newsam, S.: Motion-aware feature for improved video anomaly detection.
  In: British Machine Vision Conference (2019)

\end{thebibliography}
\end{document}